\pdfoutput=1

\documentclass[11pt]{article}

\usepackage{acl}

\usepackage{times}
\usepackage{latexsym}

\usepackage[T1]{fontenc}

\usepackage[utf8]{inputenc}

\usepackage{microtype}
\usepackage{amsmath}
\usepackage{enumitem}
\usepackage{booktabs}
\usepackage{multirow}
\usepackage{tabularx}
\usepackage{graphicx}
\usepackage{subcaption}
\usepackage{cleveref}
\usepackage{tablefootnote}
\usepackage{csquotes}
\MakeOuterQuote{"}

\crefformat{section}{\S#2#1#3} 
\crefformat{subsection}{\S#2#1#3}
\crefformat{subsubsection}{\S#2#1#3}

\newcommand{\comment}[1]{}
\newcommand{\PreserveBackslash}[1]{\let\temp=\\#1\let\\=\temp}
\newcolumntype{C}[1]{>{\PreserveBackslash\centering}p{#1}}
\newcolumntype{R}[1]{>{\PreserveBackslash\raggedleft}p{#1}}
\newcolumntype{L}[1]{>{\PreserveBackslash\raggedright}p{#1}}
\DeclareMathOperator*{\argmax}{arg\,max}

\title{End-to-End Neural Discourse Deixis Resolution in Dialogue}

\author{Shengjie Li \and Vincent Ng \\
 Human Language Technology Research Institute \\
  University of Texas at Dallas \\
  Richardson, TX 75083-0688 \\
  {\tt \{sxl180006,vince\}@hlt.utdallas.edu} \\
}

\begin{document}
\maketitle
\begin{abstract}
We adapt Lee et al.'s~\shortcite{lee-etal-2018-higher} %
span-based entity coreference model to the task of end-to-end discourse deixis resolution in dialogue, specifically by proposing extensions to their model that exploit task-specific characteristics. The resulting model, {\tt dd-utt}, achieves state-of-the-art results on the four datasets in the CODI-CRAC 2021 shared task. 
\end{abstract}

\section{Introduction}

Discourse deixis (DD) resolution, also known as abstract anaphora resolution, is an under-investigated task that involves resolving a deictic anaphor to its antecedent.
A {\em deixis} is a reference to a discourse entity such as a proposition, a description, an event, or a speech act \cite{webber:1991}. 
DD resolution is arguably more challenging than the extensively-investigated entity coreference resolution task. Recall that in entity coreference, the goal is to cluster the entity mentions in narrative text or dialogue, which are composed of pronouns, names, and nominals, so that the mentions in each cluster refer to the same real-world entity. Lexical overlap is a strong indicator of entity coreference, both among names (e.g., "President Biden", "Joe Biden") and in the resolution of nominals (e.g., linking "the president" to "President Biden"). DD resolution, on the other hand, can be viewed as a generalized case of event coreference involving the clustering of deictic anaphors, which can be pronouns or nominals, and clauses, such that the mentions in each cluster refer to the same real-world proposition/event/speech act, etc. The first example in Figure~\ref{fig:example} is an example of DD resolution in which the deictic anaphor "the move" refers to Salomon's act of issuing warrants on shares described in the preceding sentence. %
DD resolution is potentially more challenging than entity coreference resolution because (1) DD resolution involves understanding {\em clause} semantics since antecedents are clauses, and clause semantics are arguably harder to encode than noun phrase semantics; and (2) string matching plays little role in DD resolution, unlike in entity coreference.

\begin{figure}
\begin{small}
\hrule \vspace{3pt}
\textbf{Salomon Brothers International Ltd. announced it will issue warrants on shares of Hong Kong Telecommunications Ltd.} {\em The move} closely follows a similar offer by Salomon of warrants for shares of Hongkong \& Shanghai Banking Corp.

\vspace{3pt} \hrule \vspace{3pt}
 A: Would you donate to Save the Children?

 B: {\bf Yes, I will do \$10 to both.} %

 B: I am %
 of a tight budget, but I do make room for good causes.

 A: Thank you very much.

 A: The children will appreciate {\em it}. %
\vspace{3pt} \hrule
 \vspace{2mm}
     \caption{Examples of discourse deixis resolution. In each example, the deictic anaphor is italicized and the antecedent is boldfaced.}
     \label{fig:example}
     \end{small}
     \vspace{-8mm}
 \end{figure}

 \vspace{-5pt}
 
In this paper, we focus on end-to-end DD resolution in dialogue. 
The second example in Figure~\ref{fig:example} shows a dialogue between A and B in which the deictic anaphor ``it'' refers to the utterance by B in which s/he said s/he would donate \$10. %
While the deictic anaphors in dialogue are also composed of pronouns and nominals, the proportion of pronominal deictic anaphors in dialogue is much higher than that in narrative text. For instance, while 76\% of the deictic anaphors in two text corpora (ARRAU RST and GNOME) are pronominal, the corresponding percentage rises to 93\% when estimated based on seven dialogue corpora (TRAINS91, TRAINS93, PEAR, and the four CODI-CRAC 2021 development sets).
In fact, the three pronouns "that", "this", and "it" alone comprise 89\% 
of the deictic anaphors in these dialogue corpora.
The higher proportion of pronominal deictic anaphors potentially makes 
DD resolution in dialogue more challenging than those in text: since a pronoun is semantically empty, the successful resolution of a pronominal deictic anaphor depends entirely on proper understanding of its context.
In addition, it also makes DD {\em recognition} more challenging in dialogue.
For instance, while the head of a non-pronominal phrase can often be exploited to determine whether it is a deictic anaphor (e.g., "the man" cannot be a deictic anaphor, but "the move" can), such cues are absent in pronouns.

Since DD resolution can be cast as a generalized case of event coreference, a natural question is: how successful would a state-of-the-art entity coreference model be when applied to DD resolution? Recently, \newcite{kobayashi-etal-2021-neural} have applied
Xu and Choi's~\shortcite{xu-choi-2020-revealing} re-implementation of Lee et al.'s span-based entity coreference model to resolve the deictic anaphors in the DD track of the CODI-CRAC 2021 shared task after augmenting it with a {\em type prediction} model (see Section~4). Not only did they achieve the highest score on each dataset, but they beat the second-best system \cite{anikina-etal-2021-anaphora}, which is a non-span-based neural approach combined with hand-crafted rules, by a large margin. These results suggest that a span-based approach to DD resolution holds promise.

Our contributions in this paper are three-fold. First, we investigate whether {\em task-specific} observations can be exploited to extend a span-based model originally developed for entity coreference to improve its performance for end-to-end DD resolution in dialogue. 
Second, our extensions %
are effective in improving model performance, allowing our model to achieve state-of-the-art results on the CODI-CRAC 2021 shared task datasets. 
Finally, we present an empirical analysis of our model, which, to our knowledge, is the first  analysis of a state-of-the-art span-based DD resolver.

\section{Related Work}

Broadly, existing approaches to DD resolution can be divided into three categories, 
as described below.

\vspace{-2mm}
\paragraph{Rule-based approaches.} Early systems that resolve deictic expressions  are rule-based \citep{Eckert2000DialogueAS, byron-2002-resolving, navarretta-2000-abstract}. Specifically, they use predefined rules to extract anaphoric mentions, and select antecedent for each extracted anaphor based on the dialogue act types of each candidate antecedent.

\vspace{-2mm}
\paragraph{Non-neural learning-based approaches.} Early non-neural learning-based approaches to DD resolution use hand-crafted feature vectors to represent mentions \citep{strube-muller-2003-machine, Mller2008FullyAR}. A classifier is then trained to determine whether a pair of mentions is a valid antecedent-anaphor pair. 

\vspace{-2mm}
\paragraph{Deep learning-based approaches.} %
Deep learning has been applied to DD resolution. For instance, \citet{marasovic-etal-2017-mention} and \citet{anikina-etal-2021-anaphora} use a Siamese neural network, which takes as input the embeddings of two sentences, one containing a deictic anaphor and the other a candidate antecedent, to %
score %
each candidate antecedent and subsequently rank the candidate antecedents based on these scores. 
In addition, motivated by the recent successes of Transformer-based approaches to entity coreference %
(e.g., \citet{kantor-globerson-2019-coreference}),
\citet{kobayashi-etal-2021-neural} have recently proposed a 
Transformer-based approach to DD resolution, which is an end-to-end coreference system based on SpanBERT \citep{joshi-etal-2019-bert, joshi-etal-2020-spanbert}. %
Their model jointly learns mention extraction and DD resolution and has achieved state-of-the-art results in the CODI-CRAC 2021 shared task.

\section{Corpora}

We use the DD-annotated corpora provided as part of the CODI-CRAC 2021 shared task.
For training, we use the official training corpus from the shared task \citep{khosla-etal-2021-codi},  ARRAU  \citep{poesio-artstein-2008-anaphoric}, 
which consists of three conversational sub-corpora (TRAINS-93, TRAINS-91, PEAR) and two non-dialogue sub-corpora (GNOME, RST).
For validation and evaluation, we use the official development sets and test sets from the shared task. The shared task corpus is composed of %
four well-known conversational datasets: AMI \citep{McCowan2005TheAM}, LIGHT \citep{urbanek-etal-2019-learning}, Persuasion \citep{wang-etal-2019-persuasion}, and Switchboard \citep{switchboard}. %
Statistics on these corpora are provided in Table~\ref{tab:add-stats}.

\begin{table*}[t!]
\centering
\small
\begin{tabular}{@{}lllllllllll@{}}
\toprule
 &  & Total & Total & Total & Avg. & Avg. \#toks & Avg. & Avg. & Avg. & Avg. \#speakers \\ 
 &  & \#docs & \#sents & \#turns & \#sents & per sent & \#turns & \#ana & \#ante & per doc \\ \midrule
ARRAU & train & 552 & 22406 & - & 40.6 & 15.5 & - & 2.9 & 4.8 & - \\ \midrule
LIGHT & dev & 20 & 908 & 280 & 45.4 & 12.7 & 14.0 & 3.1 & 4.2 & 2.0 \\
 & test & 21 & 923 & 294 & 44.0 & 12.8 & 14.0 & 3.8 & 4.6 & 2.0 \\ \midrule
AMI & dev & 7 & 4139 & 2828 & 591.3 & 8.2 & 404.0 & 32.9 & 42.0 & 4.0 \\
 & test & 3 & 1967 & 1463 & 655.7 & 9.3 & 487.7 & 39.3 & 47.3 & 4.0 \\ \midrule
Pers. & dev & 21 & 812 & 431 & 38.7 & 11.3 & 20.5 & 4.5 & 4.5 & 2.0 \\
 & test & 28 & 1139 & 569 & 40.7 & 11.1 & 20.3 & 4.4 & 4.8 & 2.0 \\ \midrule
Swbd. & dev & 11 & 1342 & 715 & 122.0 & 11.2 & 65.0 & 11.5 & 15.9 & 2.0 \\
 & test & 22 & 3652 & 1996 & 166.0 & 9.6 & 90.7 & 12.0 & 14.7 & 2.0 \\ \bottomrule
\end{tabular}
\caption{Statistics on the datasets.} 
\label{tab:add-stats}
\end{table*}

\section{Baseline Systems}

We employ three baseline systems.

The first baseline, {\tt coref-hoi}, is Xu and Choi's~\shortcite{xu-choi-2020-revealing} re-implementation of Lee et al.'s~\shortcite{lee-etal-2018-higher} widely-used end-to-end entity coreference model. The model ranks all text spans of up to a predefined length based on how likely they correspond to entity mentions. 
For each top-ranked span $x$, the model learns a distribution $P(y)$ over its antecedents $y \in \mathcal{Y}(x)$, where $\mathcal{Y}(x)$ includes a dummy antecedent $\epsilon$ and every preceding span: 
\begin{align*}
    P(y) = \frac{e^{s(x,y)}}{\sum_{y' \in \mathcal{Y}(x)} e^{s(x,y')}}
\end{align*}

{\noindent where $s(x,y)$ is a pairwise score that incorporates two types of scores: (1) $s_m(\cdot)$, which indicates how likely a span is a mention, and (2) $s_c(\cdot)$ and $s_a(\cdot)$, which indicate how likely two spans refer to the same entity%
\footnote{See \newcite{lee-etal-2018-higher} for a description of the differences between $s_c(\cdot)$ and $s_a(\cdot)$,}
($s_a(x, \epsilon) = 0$ for dummy antecedents):}

\vspace{-4mm}
{\small
\begin{align}
    s(x,y) &= s_m(x) + s_m(y) + s_c(x,y) + s_a(x,y) \label{eq1} \\
    s_m(x) &= \texttt{FFNN}_m(g_x) \label{eq2} \\
    s_c(x,y) &= g_x^\top W_c g_y \label{eq3} \\
    s_a(x,y) &= \texttt{FFNN}_c(g_x, g_y, g_x \circ g_y, \phi(x, y)) \label{eq4}
\end{align}}
\vspace{-5mm}

{\noindent where $g_x$ and $g_y$ are the vector representations of $x$ and $y$, $W_c$ is a learned weight matrix for bilinear scoring, \texttt{FFNN($\cdot$)} is a feedforward neural network, and $\phi(\cdot)$ encodes features. Two features are used, one encoding speaker information and the other the segment distance between two spans.}

The second baseline, {\tt UTD\_NLP}%
\footnote{For an analysis of this and other resolvers competing in the CODI-CRAC 2021 shared task, see \newcite{li-etal-2021-codi}.}, is the top-performing system in the DD track of the CODI-CRAC 2021 shared task \cite{kobayashi-etal-2021-neural}. It extends {\tt coref-hoi} with a set of modifications.
Two of the most important modifications are: (1) the addition of a sentence distance feature to $\phi(\cdot)$, and (2) the incorporation into {\tt coref-hoi} a {\em type prediction} model, which predicts the type of a span. The possible types of a span $i$ are: \textsc{Antecedent} (if $i$ corresponds to an antecedent), \textsc{Anaphor} (if $i$ corresponds to an anaphor), and \textsc{Null} (if it is neither an antecedent nor an anaphor). The types predicted by the model are then used by {\tt coref-hoi} as follows: only spans predicted as \textsc{Anaphor} can be resolved, and they can only be resolved to spans predicted as \textsc{Antecedent}. %
Details of how the type prediction model is trained can be found in \newcite{kobayashi-etal-2021-neural}.

The third baseline, {\tt coref-hoi-utt}, is essentially the first baseline except that we restrict the candidate antecedents to be {\em utterances}. This restriction is motivated by the observation that the antecedents of the deictic anaphors in the CODI-CRAC 2021 datasets are all utterances. To see what an utterance is, consider again the second example in Figure~\ref{fig:example}. Each line in this dialogue is an utterance. As can be seen, an utterance roughly corresponds to a sentence, although it can also be a text fragment or simply an interjection (e.g., "uhhh"). While by definition the antecedent of a deictic anaphor can be any clause, the human annotators of the DD track of the CODI-CRAC 2021 shared task decided to restrict the unit of annotation to utterances because (1) based on previous experience it was difficult to achieve high inter-annotator agreement when clauses are used as the annotation unit \cite{poesio-artstein-2008-anaphoric}; and %
(2) unlike the sentences in narrative text, which can be composed of multiple clauses and therefore can be long, the utterances in these datasets are relatively short and can reliably be used as annotation units. From a modeling perspective, restricting candidate antecedents also has advantages. First, it substantially reduces the number of candidate antecedents and therefore memory usage, allowing our full model to fit into memory. Second, it allows us to focus on {\em resolution} rather than {\em recognition} of deictic anaphors, as the recognition of clausal antecedents remains a challenging task, especially since existing datasets for DD resolution are relatively small compared to those available for entity coreference (e.g., OntoNotes \cite{hovy-etal-2006-ontonotes}).

\section{Model}

Next, we describe our resolver, {\tt dd-utt}, which augments
{\tt coref-hoi-utt} with 10 extensions.

\vspace{-2mm}
\paragraph{E1.\ Modeling recency.} Unlike in entity coreference, where two coreferent names (e.g., "Joe Biden", "President Biden") can be far apart from each other in the corresponding document (because names are non-anaphoric), the distance between a deictic anaphor and its antecedent is comparatively smaller. To model recency, we restrict the set of candidate antecedents of an anaphor to be the utterance containing the anaphor as well as the preceding 10 utterances, the choice of which is based on our observation of the development data, where the 10 closest utterances already cover 96--99\% of the antecedent-anaphor pairs. 

\vspace{-2mm}
\paragraph{E2.\ Modeling distance.} While the previous extension allows us to restrict our attention to candidate antecedents that are close to the anaphor, it does not model the fact that the likelihood of being the correct antecedent tends to increase as its distance from the anaphor decreases. To model this relationship, we subtract the term $\gamma_1 Dist(x,y)$ from $s(x,y)$ (see Equation~(\ref{eq1})), %
where $Dist(x,y)$ is the utterance distance between anaphor $x$ and candidate antecedent $y$ and $\gamma_1$ is a tunable parameter that controls the importance of utterance distance in the resolution process. Since $s(x,y)$ is used to rank candidate antecedents, modeling utterance distance by updating $s(x,y)$ will allow distance to have a direct impact on DD resolution.

\vspace{-2mm}
\paragraph{E3.\ Modeling candidate antecedent length.} Some utterances are pragmatic in nature and do not convey any important information. Therefore, they cannot serve as antecedents of deictic anaphors. Examples include "Umm", "Ahhhh... okay", "that's right", and "I agree". Ideally, the model can identify such utterances and prevent them from being selected as antecedents. We hypothesize that we could help the model by modeling such utterances. To do so, we observe that such utterances tend to be short and model them by penalizing shorter utterances. Specifically, we subtract the term $\gamma_2 \frac{1}{Length(y)}$ from $s(x,y)$, where $Length(y)$ is the number of words in candidate antecedent $y$ and $\gamma_2$ is a tunable parameter that controls the importance of candidate antecedent length in resolution. %

\vspace{-2mm}
\paragraph{E4.\ Extracting candidate anaphors.}
As mentioned before, the deictic anaphors in dialogue are largely composed of pronouns.
Specifically, in our development sets, the three pronouns "that", "this", and `it' 
alone account for %
74--88\% of the anaphors. 
Consequently, we extract candidate deictic anaphors as follows:
instead of allowing each span of length $n$ or less to be a candidate anaphor, we only allow a span to be a candidate anaphor if its underlying word/phrase has appeared at least once in the training set as a deictic anaphor. 

\vspace{-2mm}
\paragraph{E5.\ Predicting anaphors.}
Now that we have the candidate anaphors, our next extension involves predicting which of them are indeed deictic anaphors. To do so, we retrain the type prediction model in {\tt UTD\_NLP}, which is a FFNN that takes as input the (contextualized) span representation $g_i$ of candidate anaphor $i$ and outputs a vector $ot_i$ of dimension 2 in which the first element denotes the likelihood that $i$ is a deictic anaphor and the second element denotes the likelihood that $i$ is not a deictic anaphor. $i$ is predicted as a deictic anaphor if and only if the value of the first element of $ot_i$ is bigger than its second value:

\vspace{-6mm}
\begin{align*}
    ot_i &= \texttt{FFNN}(g_i) \\
    t_i &= \argmax_{x \in \{\text{A}, \text{NA}\}} ot_i(x)
\end{align*}

\vspace{-1mm}
\noindent{where \text{A} ({\sc Anaphor}) and \text{NA} ({\sc Non-Anaphor}) are the two possible types. Following {\tt UTD\_NLP}, this type prediction model is jointly trained with the resolution model. Specifically, we compute the cross-entropy loss using $ot_i$, multiply it by a type loss coefficient $\lambda$, and add it to the loss function of {\tt coref-hoi-utt}. $\lambda$ is a tunable parameter that controls the importance of type prediction relative to DD resolution.}

\vspace{-2mm}
\paragraph{E6.\ Modeling the relationship between anaphor recognition and resolution.}
In principle, the model should resolve a candidate anaphor to a non-dummy candidate antecedent if it is predicted to be a deictic anaphor by the type prediction model. However, type prediction is not perfect, and enforcing this consistency constraint, which we will refer to as {\bf C1}, will allow errors in type prediction to be propagated to DD resolution. For example, if a non-deictic anaphor is misclassified by the type prediction model, then it will be (incorrectly) resolved to a non-dummy antecedent. To alleviate error propagation, we instead %
enforce {\bf C1} in a {\em soft} manner. To do so, we define a penalty function $p_1$, which imposes a penalty on span $i$ if {\bf C1} is violated (i.e., a deictic anaphor is resolved to the dummy antecedent), as shown below:
\begin{equation*}
\resizebox{\hsize}{!}{$
p_1(i)=
\begin{cases}
0 & \argmax\limits_{y_{is}\in \mathcal{Y}} t_i = \text{NA} \\
ot_i(\text{A}) - ot_i(\text{NA}) & \text{otherwise}
\end{cases}$}
\end{equation*}
Intuitively, $p_1$ estimates the minimum amount %
to be adjusted so that span $i$’s type is not {\sc Anaphor}.

We incorporate $p_1$ into the model as a penalty term in $s$ (Equation~(\ref{eq1})). Specifically, we redefine $s(i,j)$ when $j = \epsilon$, as shown below: 
\begin{equation*}
s(i, \epsilon) = s(i, \epsilon) - [\gamma_3 p_1(i)]
\end{equation*}
where $\gamma_3$ is a positive constant that controls the hardness of {\bf C1}. The smaller $\gamma_3$ is, the softer {\bf C1} is. Intuitively, if {\bf C1} is violated, $s(i, \epsilon)$ will be lowered by the penalty term, and the dummy antecedent will less likely be selected as the antecedent of $i$.

\vspace{-2mm}
\paragraph{E7.\ Modeling the relationship between non-anaphor recognition and resolution.}
Another consistency constraint that should be enforced is that the model should resolve a candidate anaphor to the dummy antecedent if it is predicted as a non-deictic anaphor by the type prediction model. As in Extension~E6, we will enforce this constraint, which we will refer to as {\bf C2}, in a soft manner by defining a penalty function $p_2$, as shown below:
\begin{equation*}
\resizebox{\hsize}{!}{$
p_2(i)=
\begin{cases}
ot_i(\text{NA}) - ot_i(\text{A}) & \argmax\limits_{y_{is}\in \mathcal{Y}} t_i = \text{NA} \\
0 & \text{otherwise}
\end{cases}$}
\end{equation*}

{\noindent Then we redefine $s(i,j)$ when $j \neq \epsilon$ as follows: }
\begin{equation*}
s(i, j) = s(i, j) - [\gamma_4 p_1(i)]
\end{equation*}
where $\gamma_4$ is a positive constant that controls the hardness of {\bf C2}. 
Intuitively, if {\bf C2} is violated, $s(i, j)$ will be lowered by the penalty term, and $j$ will less likely be selected as the antecedent of $i$.

\vspace{-2mm}
\paragraph{E8.\ Encoding candidate anaphor context.}
Examining Equation (\ref{eq1}), we see that $s(x,y)$ is computed based on the span representations of $x$ and $y$. While these span representations are contextualized, the contextual information they encode is arguably limited. As noted before, most of the deictic anaphors in dialogue are pronouns, which are semantically empty. As a result, we hypothesize that we could improve the resolution of these deictic anaphors if we explicitly modeled their contexts. Specifically, we represent the context of a candidate anaphor using the embedding of the utterance in which it appears and add the resulting embedding as features to both the bilinear score $s_c(x,y)$ and the concatenation-based score $s_a(x,y)$:
\begin{align*}
    s_c(x,y) &= g_x^\top W_c g_y + g_s^\top W_s g_y\\
    s_a(x,y) &= \texttt{FFNN}_c(g_x, g_y, g_x \circ g_y, g_{s}, \phi(x, y))
\end{align*}
where $W_c$ and $W_s$ are learned weight matrices, $g_s$ is the embedding of the utterance $s$ in which candidate anaphor $x$ appears, and $\phi(x, y)$ encodes the relationship between $x$ and $y$ as features.

\begin{table}[t!]
\begin{center}
    \begin{small}
    \begin{tabular}{p{7.2cm}} \toprule
\multicolumn{1}{c}{\bf Filling words} \\
yeah,  okay,  ok,  uh,  right,  so,  hmm,  well,  um,  oh,  mm,  yep,  hi,  ah,  whoops,  alright,  shhhh,  yes,  ay,  hello,  aww,  alas,  ye,  aye,  uh-huh,  huh,  wow,  www,  no,  and,  but,  again,  wonderful,  exactly,  absolutely,  actually,  sure thanks,  awesome,  gosh,  ooops \\ \midrule
 \multicolumn{1}{c}{\bf Reporting verbs} \\
 command,  mention,  demand,  request,  reveal,  believe,  guarantee,  guess,  insist,  complain,  doubt,  estimate,  warn,  learn,  realise,  persuade,  propose,  announce,  advise,  imagine,  boast,  suggest,  remember,  claim,  describe,  see,  understand,  discover,  answer,  wonder,  recommend,  beg,  prefer,  suppose,  comment,  think,  argue,  consider,  swear,  ask,  agree,  explain,  report,  know,  tell,  decide,  discuss,  repeat,  invite,  reply,  expect,  forget,  add,  fear,  hope,  say,  feel,  observe,  remark,  confirm,  threaten,  teach,  forbid,  admit,  promise,  deny,  state,  mean,  instruct \\ \bottomrule
\end{tabular}
\end{small}
\end{center}
\caption{Lists of filtered words.}
\label{tab:FilteredWords}
\vspace{-2mm}
\end{table}

\vspace{-2mm}
\paragraph{E9.\ Encoding the relationship between candidate anaphors and antecedents.}
As noted in Extension~E8, $\phi(x,y)$ encodes the relationship between candidate anaphor $x$ and candidate antecedent $y$. In {\tt UTD\_NLP}, $\phi(x,y)$ is composed of three features, including two features from {\tt coref-hoi-utt} (i.e., the speaker id and the {\em segment} distance between $x$ and $y$) and one feature that encodes the {\em utterance} distance between them. Similar to the previous extension, we hypothesize that we could better encode the relationship between $x$ and $y$ using additional features.
Specifically, we incorporate an additional feature into $\phi(x,y)$ that encodes the utterance distance between $x$ and $y$. Unlike the one used in {\tt UTD\_NLP}, this feature aims to more accurately capture proximity by ignoring {\em unimportant} sentences (i.e., those that  contain only interjections, filling words, reporting verbs, and punctuation) when computing utterance distance. The complete list of filling words and reporting verbs that we filter can be found in Table~\ref{tab:FilteredWords}.

\vspace{-2mm}
\paragraph{E10.\ Encoding candidate antecedents.}
In {\tt coref-hoi-utt}, a candidate antecedent is simply encoded using its span representation. We hypothesize that we could better encode a candidate antecedent using additional {\em features}. Specifically, we employ seven features to encode a candidate antecedent $y$ and incorporate them into $\phi(x,y)$:  (1) the number of words in $y$; (2) the number of nouns  in $y$; (3) the number of verbs  in $y$; (4) the number of adjectives in $y$; (5) the number of content word overlaps between $y$ and the portion of the utterance containing the anaphor that precedes the anaphor; (6) whether $y$ is the longest among the candidate antecedents; and (7) whether $y$ has the largest number of content word overlap (as computed in Feature \#5) among the candidate antecedents. Like Extension~E3, some features implicitly encode the length of a candidate antecedent. Despite this redundancy, we 
believe the redundant information could be exploited by the model differently and may therefore have varying degrees of impact on it.

\section{Evaluation}

\subsection{Experimental Setup}

\begin{table*}[]
\centering
\small
\begin{tabular}{@{}lcccccccccr@{}}
\toprule
 & \multicolumn{5}{c}{Resolution} & \multicolumn{5}{c}{Recognition} \\ 
 & LIGHT & AMI & Pers. & Swbd. & Avg. & LIGHT & AMI & Pers. & Swbd. & Avg. \\ 
 \cmidrule(r){1-1} \cmidrule(lr){2-6} \cmidrule(l){7-11}
UTD\_NLP & 42.7 & 35.4 & 39.6 & 35.4 & 38.3 & 70.1 & 61.0 & 69.9 & 68.1 & 67.3 \\
coref-hoi & 42.7 & 30.7 & 49.7 & 35.4 & 39.6 & 70.9 & 49.3 & 67.8 & 61.9 & 62.5 \\
coref-hoi-utt & 42.3 & 35.0 & 53.3 & 34.1 & 41.2 & 70.3 & 52.4 & 71.0 & 60.6 & 63.6 \\
dd-utt & 48.2 & 43.5 & 54.9 & 47.2 & 48.5 & 71.3 & 56.9 & 71.4 & 65.2 & 66.2 \\ \bottomrule
\end{tabular}
\caption{Resolution and recognition results on the four test sets.}
\label{tab:results}
\vspace{-4pt}
\end{table*}

\paragraph{Evaluation metrics.}
We obtain the results of DD resolution using the Universal Anaphora Scorer 
\cite{yu-etal-2022-universal}. Since DD resolution is viewed as a generalized case of event coreference, the scorer reports performance in terms of CoNLL score, which is the unweighted average of the F-scores of three 
coreference scoring metrics, namely MUC \citep{vilain-etal-1995-model}, B$^3$ \citep{Bagga98algorithmsfor}, and CEAF$_e$ \citep{luo-2005-coreference}.
In addition, we report the results of deictic anaphor recognition. We express recognition results in terms of Precision (P), Recall (R) and F-score, considering an anaphor correctly recognized if it has an exact match with a gold anaphor in terms of boundary.

\begin{table}[t]
\centering
\small
\begin{tabular}{ccccc}
\toprule
 & LIGHT & AMI & Pers. & Swbd. \\ \midrule
Type loss coefficient $\lambda$ & 800 & 800 & 800 & 800 \\
$\gamma_1$ & 1 & 1 & 1 & 1 \\
$\gamma_2$ & 1 & 1 & 1 & 1 \\
$\gamma_3$ & 5 & 10 & 10 & 5 \\
$\gamma_4$ & 5 & 5 & 5 & 5 \\ \bottomrule
\end{tabular}
\caption{Parameter values enabling {\tt dd-utt} to achieve the best CoNLL score on each development set.}
\label{tab:params}
\vspace{-4pt}
\end{table}

\vspace{-2mm}
\paragraph{Model training and parameter tuning.} For \texttt{coref-hoi} and \texttt{coref-hoi-utt}, we use SpanBERT$_{\text{Large}}$ as the encoder and
reuse the hyperparameters from \citet{xu-choi-2020-revealing} with the only exception of the maximum span width: for \texttt{coref-hoi}, we increase the maximum span width from 30 to 45 in order to cover more than 97\% of the antecedent spans; for \texttt{coref-hoi-utt} we use 15 as the maximum span width, which covers more than 99\% of the anaphor spans in the training sets.
For \texttt{UTD\_NLP}, we simply take 
the outputs produced by the model on the test sets and report the results obtained by running the scorer on the outputs.%
\footnote{Since the shared task participants were allowed to submit their system outputs multiple times to the server to obtain results on the test sets, {\tt UTD\_NLP}'s results could be viewed as results obtained by tuning parameters on the test sets.}
For {\tt dd-utt}, we use SpanBERT$_{\text{Large}}$ as the encoder. 
Since we do not rely on span enumerate to generate candidate spans, the maximum span width can be set to any arbitrary number that is large enough to cover all candidate antecedents and anaphors. In our case, we use 300 as our maximum span width. 
We tune the parameters (i.e., $\lambda$, $\gamma_1$, $\gamma_2$, $\gamma_3$, $\gamma_4$) 
using grid search to maximize CoNLL score on development data. 
For the type loss coefficient, we search out of \{0.2, 0.5, 1, 200, 500, 800, 1200, 1600\}, and for $\gamma$, we search out of \{1, 5, 10\}. 
We reuse other hyperparameters from \citet{xu-choi-2020-revealing}. 

All models %
are trained for 30 epochs with a dropout rate of 0.3 and early stopping. 
We use $1 \times 10^{-5}$ as our BERT learning rate and $3 \times 10^{-4}$ as our task learning rate. Each experiment is run using a random seed of 11 %
and takes less than three hours to train on an NVIDIA RTX A6000 48GB. 

\vspace{-2mm}
\paragraph{Train-dev partition.}
Since we have four test sets, %
we use ARRAU and all dev sets other than the one to be evaluated on for model training %
and the remaining dev set for parameter tuning. For example, when evaluating on AMI$_{\text{test}}$, we train models on ARRAU, LIGHT$_{\text{dev}}$, Persuasion$_{\text{dev}}$ and Switchboard$_{\text{dev}}$ and use AMI$_{\text{dev}}$ for tuning.

\subsection{Results}

Recall that our goal is to perform end-to-end DD resolution, which corresponds to the Predicted evaluation setting in the CODI-CRAC  shared task.

\vspace{-2mm}
\paragraph{Overall performance.} Recognition results (expressed in F-score) and resolution results (expressed in CoNLL score) of the three baselines and our model on the four test sets are shown in Table \ref{tab:results}, where the Avg.\ columns report the macro-averages of the corresponding results on the four test sets, and the parameter settings 
that enable our model to achieve the highest CoNLL scores on the development sets are shown in Table \ref{tab:params}. 
Since {\tt coref-hoi} and {\tt coref-hoi-utt} do not explicitly identify deictic anaphors, we assume that all but the first mentions in each output cluster are anaphors when computing recognition precision; and while {\tt UTD\_NLP} (the top-performing system in the shared task) does recognize anaphors, we still make the same assumption when computing its recognition precision since the anaphors are not explicitly marked in the output (recall that we computed results of {\tt UTD\_NLP} based on its outputs).

\begin{table*}[]
\centering
\small
\centering
\begin{tabular}{cccccccccc}
\toprule
& &  \multicolumn{4}{c}{Resolution} & \multicolumn{4}{c}{Recognition} \\ 
anaphor &  count & UTD\_NLP & coref-hoi & coref-hoi-utt & dd-utt & UTD\_NLP & coref-hoi & coref-hoi-utt & dd-utt \\ 
\cmidrule(r){1-1} \cmidrule(lr){2-2} \cmidrule(lr){3-6} \cmidrule(l){7-10}
that &  402 & 48.7 & 49.1 & 50.4 & 60.7 & 79.1 & 72.7 & 72.9 & 76.8 \\
it &  95 & 21.8 & 27.5 & 28.5 & 37.0 & 36.5 & 36.4 & 37.0 & 35.5 \\
this &  25 & 20.7 & 25.1 & 26.9 & 25.8 & 51.7 & 49.2 & 49.3 & 56.5 \\
which &  10 & 4.5 & 8.2 & 14.9 & 14.4 & 33.3 & 28.6 & 42.1 & 35.3 \\
Others &  52 & 33.0 & 39.6 & 41.2 & 48.5 & 33.8 & 28.1 & 40.0 & 10.3 \\ \bottomrule
\end{tabular}
\caption{Per-anaphor recognition and resolution results on the test sets.}
\label{tab:per-anaphor}
\end{table*}
\vspace{-4pt}

We test the statistical significance among the four models using two-tailed Approximate Randomization \cite{noreen:book}. For recognition, the models are statistically indistinguishable from each other w.r.t.\ their Avg.\ score ($p < 0.05$). For resolution, {\tt dd-utt} is highly significantly better than the baselines w.r.t.\ Avg.\ ($p < 0.001$), while the three baselines 
are statistically indistinguishable from each other. These results suggest that (1) {\tt dd-utt}'s superior resolution performance stems from better antecedent selection, not better anaphor recognition; and (2) the restriction of candidate antecedents to utterances in {\tt coref-hoi-utt} does not enable the resolver to yield significantly better resolution results than {\tt coref-hoi}.

\vspace{-2mm}
\paragraph{Per-anaphor results.} Next, we show the recognition and resolution results of the four models on the most frequently occurring deictic anaphors in Table~\ref{tab:per-anaphor} after micro-averaging them over the four test sets. Not surprisingly, "that" is the most frequent deictic anaphor on the test sets, appearing as an anaphor 402 times on the test sets and contributing to 68.8\% of the anaphors. This is followed by "it" (16.3\%) and "this" (4.3\%). Only 8.9\% of the anaphors are not among the top four anaphors. 

Consider first the recognition results. As can be seen, "that" has the highest recognition F-score among the top anaphors. This is perhaps not surprising given the comparatively larger number of "that" examples the models are trained on. While "it" occurs more frequently than "this" as a deictic anaphor, its recognition performance is lower than that of "this". This is not surprising either: "this", when used as a pronoun, is more likely to be deictic than "it", although both of them can serve as a coreference anaphor and a bridging anaphor. In other words, it is comparatively more difficult to determine whether a particular occurrence of "it" is deictic. Overall, UTD\_NLP recognizes more anaphors than the other models.

Next, consider the resolution results. To obtain the CoNLL scores for a given anaphor, we retain all and only those clusters containing the anaphor in both the gold partition and the system partition and apply the official scorer to them. Generally, the more frequently occurring an anaphor is, the better its resolution performance is. Interestingly, for the "Others" category, {\tt dd-utt} achieves the highest resolution results despite having the lowest recognition performance. In contrast, while {\tt UTD\_NLP} achieves the best recognition performance on average, its resolution results are among the worst.

\vspace{-2mm}
\paragraph{Per-distance results.} 
To better understand how resolution results vary with the utterance distance between a deictic anaphor and its antecedent, we show in Table~\ref{tab:dist} the number of correct and incorrect links predicted by the four models for each utterance distance on the test sets. For comparison purposes, we show at the top of the table the distribution of gold links over utterance distances.
Note that a distance of 0 implies that the anaphor refers to the utterance in which it appears.

A few points deserve mention. First, the distribution of gold links is consistent with our
intuition: a deictic anaphor most likely has the immediately preceding utterance (i.e., distance =
1) as its referent. In addition, the number of links drops as distance increases, and more than 90\% of the antecedents are at most four utterances away from their anaphors. 
Second, although {\tt UTD\_NLP} recognizes more anaphors than the other models, it is the most conservative w.r.t.\ link identification, predicting the smallest number of correct and incorrect links for almost all of the utterance distances. 
Third, %
{\tt dd-utt} is better than the other models %
at (1) identifying short-distance anaphoric dependencies, particularly when distance $\leq 1$, and 
(2) positing fewer erroneous long-distance anaphoric dependencies. These results provide suggestive evidence of {\tt dd-utt}'s success at modeling recency and distance explicitly. 
Finally, these results suggest that resolution difficulty increases with distance: except for {\tt UTD\_NLP}, none of the models can successfully recognize a link when distance $> 5$.

\begin{table}[t!]
\small
\centering
\begin{tabular}{@{}lccccccc@{}}
\toprule
 & 0 & 1 & 2 & 3 & 4 & 5 & \textgreater{}5 \\ 
\midrule
\multicolumn{8}{c}{Distribution of gold links over utterance distances} \\
Gold & 90 & 209 & 97 & 46 & 21 & 8 & 19 \\
\midrule
\multicolumn{8}{c}{Distribution of correctly predicted links} \\
UTD\_NLP & 28 & 64 & 23 & 9 & 4 & 2 & 1 \\
coref-hoi & 22 & 108 & 42 & 13 & 5 & 3 & 0 \\
coref-hoi-utt & 23 & 118 & 49 & 11 & 6 & 2 & 0 \\
dd-utt & 40 & 142 & 45 & 17 & 5 & 3 & 0 \\
\midrule
\multicolumn{8}{c}{Distribution of incorrectly predicted links} \\
UTD\_NLP & 16 & 56 & 31 & 24 & 9 & 6 & 52 \\
coref-hoi & 83 & 148 & 55 & 33 & 22 & 13 & 58 \\
coref-hoi-utt & 64 & 123 & 66 & 36 & 21 & 10 & 48 \\
dd-utt & 43 & 131 & 57 & 24 & 6 & 7 & 7 \\
\bottomrule
\end{tabular}
\caption{Distribution of links over the utterance distances between the anaphor and the antecedents.}
\label{tab:dist}
\end{table}
\vspace{-4pt}

\vspace{-2mm}
\paragraph{Ablation results.}
To evaluate the contribution of each extension presented in Section~5 to {\tt dd-utt}'s resolution performance, we show in Table~\ref{tab:ablation} ablation results, which we obtain by removing one extension at a time from {\tt dd-utt} and retraining it. For ease of comparison, we show in the first row of the table %
the CoNLL scores of {\tt dd-utt}.

A few points deserve mention. First, when E1 (Modeling recency) is ablated, we use as candidate antecedents the 10 highest-scoring candidate antecedents for each candidate anaphor according to 
$s_c(x,y)$ (Equation~(\ref{eq3})).
Second, when one of E2, E3, E6, and E7 is ablated, we set the corresponding $\lambda$ to zero.
Third, when E4 is ablated, candidate anaphors are extracted in the same way as in {\tt coref-hoi} and {\tt coref-hoi-utt}, where the top spans learned by the model will serve as candidate anaphors.
Fourth, when E5 is ablated, E6 and E7 will also be ablated because the penalty functions $p_1$ and $p_2$ need to be computed based on the output of the type prediction model in E5.

\begin{table}[t!]
\centering
\small
\begin{tabular}{@{}lccccc@{}}
\toprule
 & LIGHT & AMI & Pers. & Swbd. & Avg. \\ \midrule
dd-utt & 48.2 & 43.5 & 54.9 & 47.2 & 48.5 \\
$-$ E1 & 47.1 & 36.4 & 53.6 & 46.2 & 45.8 \\
$-$ E2 & 47.4 & 40.0 & 56.5 & 48.9 & 48.2 \\
$-$ E3 & 45.3 & 44.6 & 53.7 & 49.0 & 48.1 \\
$-$ E4 & 46.4 & 42.8 & 56.7 & 45.6 & 47.9 \\
$-$ E5 & 43.6 & 43.9 & 50.2 & 47.4 & 46.3 \\
$-$ E6 & 46.1 & 43.1 & 50.8 & 47.2 & 46.8 \\
$-$ E7 & 48.6 & 43.6 & 56.0 & 49.4 & 49.4 \\
$-$ E8 & 43.3 & 39.4 & 52.5 & 50.1 & 46.3 \\
$-$ E9 & 44.6 & 43.2 & 52.1 & 47.7 & 46.9 \\
$-$ E10 & 47.3 & 39.3 & 57.5 & 49.7 & 48.5 \\ %
\bottomrule
\end{tabular}
\caption{Resolution results of ablated models.}
\label{tab:ablation}
\end{table}
\vspace{-4pt}

We use two-tailed Approximate Randomization to determine which of these ablated models is statistically different from {\tt dd-utt} w.r.t.\ the Avg.\ score. Results show that except for the model in which E1 is ablated, all of the ablated models are statistically indistinguishable from {\tt dd-utt} ($p < 0.05$). Note that these results do {\em not} imply that nine of the extensions fail to contribute positively to {\tt dd-utt}'s resolution performance: it only means that none of them is useful in the presence of other extensions w.r.t.\ Avg. We speculate that  (1) some of these extensions model overlapping phenomena (e.g., both E2 and E9 model utterance distance); (2) when the model is retrained, the learner manages to adjust the network weights so as to make up for the potential loss incurred by ablating an extension; and (3) large fluctuations in performance can be observed on individual datasets in some of the experiments, but they may just disappear after averaging.
Experiments are needed to determine the reason. %

\subsection{Error Analysis}

Below we analyze the %
errors 
made by {\tt dd-utt}.

\vspace{-2mm}
\paragraph{DD anaphora recognition precision errors.}
A common type of recognition precision errors involves misclassifying a coreference anaphor as a deictic anaphor. Consider the first example in Figure \ref{fig:example-errors}, in which the pronoun "that" is a coreference anaphor with "voice recognition" as its antecedent but is misclassified as a deictic anaphor with the whole sentence as its antecedent. This type of error occurs because virtually all of the frequently occurring deictic anaphors, including "that", "it", "this", and "which", appear as a coreference anaphor in some contexts and as a deictic anaphor in other contexts, and distinguishing between the two different uses of these anaphors could be challenging.

\begin{figure}[t]
\begin{small}
\vspace{3pt} \hrule \vspace{3pt}

A: \hangindent=11pt The design should minimize R\_S\_I and be easy to locate and we were still slightly ambivalent as to whether to use {\em voice recognition} there, though {\em that} did seem to be the favored strategy, but there was also, on the sideline, the thought of maybe having a beeper function.

\vspace{3pt} \hrule \vspace{3pt}

A: {\bf Sounds like a blessed organization.}

B: Yes, {\em it} does.

\vspace{3pt} \hrule \vspace{3pt}

A: \hangindent=11pt {\bf Did you know they've won over 7 different awards for their charitable work?}

A: \hangindent=11pt {\em As a former foster kid, it makes me happy to see this place bring such awareness to the issues and needs of our young.}

B: I am not surprised to hear {\em that} at all.

\vspace{3pt} \hrule 
 \vspace{2mm}
     \caption{Examples illustrating the three majors types of errors made by {\tt dd-utt}.}
     \label{fig:example-errors}
     \end{small}
     \vspace{-8mm}
 \end{figure}

\vspace{-2mm}
\paragraph{DD anaphor recognition recall errors.}
Consider the second example in Figure \ref{fig:example-errors}, in which %
"it" is a deictic anaphor that refers to the boldfaced utterance, but {\tt dd-utt} fails to identify this and many other occurrences of "it" as deictic, probably because %
"it" is more likely to be a coreference anaphor than a deictic anaphor: in the dev sets, 80\% of the occurrences of "it" are coreference anaphors while only 5\% are deictic anaphors.

 \vspace{-2mm}
\paragraph{DD resolution precision errors.}
A major source of DD resolution precision errors can be attributed to the model's failure in properly understanding the context in which a deictic anaphor appears.
Consider the third example in Figure \ref{fig:example-errors}, in which "that" is a deictic anaphor that refers to the boldfaced utterance. While {\tt dd-utt} correctly identifies "that" as a deictic anaphor, it erroneously posits the italicized utterance as its antecedent. This example is interesting in that {\em without} looking at the boldfaced utterance, the italicized utterance is a plausible antecedent for "that" because "I am not surprised to hear that at all" can be used as a response to almost every statement.  However, when both the boldfaced utterance and the italicized utterance are taken into consideration, it is clear that the boldfaced utterance is the correct antecedent for "that" because winning over seven awards for some charitable work is certainly more surprising than seeing a place bring awareness to the needs of the young. Correctly resolving this anaphor, however, requires modeling the emotional implication of its context.

\subsection{Further Analysis}

Next, we %
analyze the deictic anaphors correctly resolved by {\tt dd-utt} but erroneously resolved by the baseline resolvers.

\begin{figure}[t!]
\begin{small}
\vspace{3pt} \hrule \vspace{3pt}
 
A: {\em You want your rating to be a two?}

A: Is that what you're saying?

B: Yeah, I just got it the other way.

B: Uh in Yep, I just got

A: Okay.

A: So, I'll work out the average for that again at the end.

A: It's very slightly altered. Okay, and we're just waiting for your rating.

B: two point five

C: {\em Its just two point five for that one.}

A: Two point five, okay.

D: Yeah.

A: {\bf Losing one decimal place, {\em that} is okay.}

\vspace{3pt} \hrule 
 \vspace{2mm}
     \caption{Example in which the correct antecedent is identified by {\tt dd-utt} but not by the baseline resolvers.}
     \label{fig:example-errors-appendix}
     \end{small}
     \vspace{-8mm}
 \end{figure}

The example shown in Figure \ref{fig:example-errors-appendix} is one such case.
In this example, {\tt dd-utt} successfully extracts the anaphor "that" and resolves it to the correct antecedent "Losing one decimal place, that is okay". 
{\tt UTD\_NLP} fails to extract "that" as a deictic anaphor. 
While {\tt coref-hoi} correctly extracts the anaphor, it incorrectly selects "You want your rating to be a two?" as the antecedent. From a cursory look at this example one could infer that this candidate antecedent is highly unlikely to be the correct antecedent since it is 10 utterances away from the anaphor. As for {\tt coref-hoi-utt}, the resolver successfully extracts the anaphor but incorrectly selects "Its just two point five for that one" as the antecedent, which, like the antecedent chosen by {\tt coref-hoi}, is farther away from the anaphor than the correct antecedent is. %
We speculate that
{\tt coref-hoi} and {\tt coref-hoi-utt} %
fail to identify the correct antecedent because they do not explicitly model distance 
and therefore may not have an idea about how far a candidate antecedent is from the anaphor under consideration. 
The additional features that {\tt dd-utt} has access to, including the features that encode sentence distance as well as those that capture contextual information, may have helped {\tt dd-utt} choose the correct antecedent,
but additional analysis is needed to determine the reason.

\section{Conclusion}

We presented an end-to-end discourse deixis resolution model that augments Lee et al.'s~\shortcite{lee-etal-2018-higher} span-based entity coreference model with 10 extensions. The resulting model achieved state-of-the-art results on the CODI-CRAC 2021 datasets. 
We employed a variant of this model in our recent participation in the discourse deixis track of the CODI-CRAC 2022 shared task \cite{yu-etal-2022-codi} and achieved the best results (see \newcite{li-etal-2022-neural-anaphora} for details).
To facilitate replicability, we make our source code publicly available.%
\footnote{See our website at \url{https://github.com/samlee946/EMNLP22-dd-utt} for our source code.}

\section*{Limitations}

Below we discuss several limitations of our work.

\vspace{-2mm}
\paragraph{Generalization to corpora with clausal antecedents.} As mentioned in the introduction, the general discourse deixis resolution task involves resolving a deictic anaphor to a clausal antecedent. The fact that our resolver can only resolve anaphors to utterances raises the question of whether it can be applied to resolve deictic anaphors in texts where antecedents can be clauses. To apply our resolver to such datasets, all we need to do is to expand the set of candidate antecedents of an anaphor to include those clauses that precede it. While corpora annotated with clausal antecedents exist (e.g., TRAINS-91 and TRAINS-93), we note that the decision made by the CODI-CRAC 2021 shared task organizers to use utterances as the unit of annotation has to do with annotation quality, as the inter-annotator agreement on the selection of clausal antecedents tends to be low \cite{poesio-artstein-2008-anaphoric}, 

\vspace{-2mm}
\paragraph{Discourse deixis resolution in dialogue vs.\ narrative text.} Whether our model will generalize well to non-dialogue datasets (e.g., narrative text) is unclear. Given the differences between dialogue and non-dialogue datasets (e.g., the percentage of pronominal anaphors), we speculate that the performance of our resolver will take a hit when applied to resolving deictic anaphors in narrative text.

\vspace{-2mm}
\paragraph{Size of training data.} We believe that the performance of our resolver is currently limited in part by the small amount of data on which it was trained. The annotated corpora available for training a discourse deictic resolver 
is much smaller than those available for training an entity coreference resolver (e.g., OntoNotes \cite{hovy-etal-2006-ontonotes}).

\vspace{-2mm}
\paragraph{Data biases.} Generally, our work should not cause any significant risks. However, language varieties not present in training data can potentially amplify existing inequalities and contribute to misunderstandings.

\section*{Acknowledgments}
We thank the three anonymous reviewers for their
insightful 
comments on an earlier
draft of the paper. 
This work was supported in part by NSF Grant IIS-1528037.
Any opinions, findings, conclusions or recommendations expressed in this paper are those of the authors and do not necessarily reflect the views or official policies, either expressed or implied, of the NSF. 

\bibliography{anthology,custom}

\begin{thebibliography}{29}
\expandafter\ifx\csname natexlab\endcsname\relax\def\natexlab#1{#1}\fi

\bibitem[{Anikina et~al.(2021)Anikina, Oguz, Skachkova, Tao, Upadhyaya, and
  Kruijff-Korbayova}]{anikina-etal-2021-anaphora}
Tatiana Anikina, Cennet Oguz, Natalia Skachkova, Siyu Tao, Sharmila Upadhyaya,
  and Ivana Kruijff-Korbayova. 2021.
\newblock \href {https://doi.org/10.18653/v1/2021.codi-sharedtask.3} {Anaphora
  resolution in dialogue: Description of the {DFKI}-{T}alking{R}obots system
  for the {CODI}-{CRAC} 2021 shared-task}.
\newblock In \emph{Proceedings of the CODI-CRAC 2021 Shared Task on Anaphora,
  Bridging, and Discourse Deixis in Dialogue}, pages 32--42, Punta Cana,
  Dominican Republic. Association for Computational Linguistics.

\bibitem[{Bagga and Baldwin(1998)}]{Bagga98algorithmsfor}
Amit Bagga and Breck Baldwin. 1998.
\newblock Algorithms for scoring coreference chains.
\newblock In \emph{Proceedings of the LREC Workshop on Linguistics
  Coreference}, pages 563--566.

\bibitem[{Byron(2002)}]{byron-2002-resolving}
Donna~K. Byron. 2002.
\newblock \href {https://doi.org/10.3115/1073083.1073099} {Resolving pronominal
  reference to abstract entities}.
\newblock In \emph{Proceedings of the 40th Annual Meeting of the Association
  for Computational Linguistics}, pages 80--87, Philadelphia, Pennsylvania,
  USA. Association for Computational Linguistics.

\bibitem[{Eckert and Strube(2000)}]{Eckert2000DialogueAS}
Miriam Eckert and Michael Strube. 2000.
\newblock Dialogue acts, synchronizing units, and anaphora resolution.
\newblock \emph{Journal of Semantics}, 17:51--89.

\bibitem[{Godfrey et~al.(1992)Godfrey, Holliman, and McDaniel}]{switchboard}
John~J. Godfrey, Edward~C. Holliman, and Jane McDaniel. 1992.
\newblock \href {https://doi.org/10.1109/ICASSP.1992.225858} {Switchboard:
  telephone speech corpus for research and development}.
\newblock In \emph{Proceedings of the 1992 IEEE International Conference on
  Acoustics, Speech, and Signal Processing (ICASSP-92)}, volume~1, pages
  517--520.

\bibitem[{Hovy et~al.(2006)Hovy, Marcus, Palmer, Ramshaw, and
  Weischedel}]{hovy-etal-2006-ontonotes}
Eduard Hovy, Mitchell Marcus, Martha Palmer, Lance Ramshaw, and Ralph
  Weischedel. 2006.
\newblock \href {https://aclanthology.org/N06-2015} {{O}nto{N}otes: The 90{\%}
  solution}.
\newblock In \emph{Proceedings of the Human Language Technology Conference of
  the {NAACL}, Companion Volume: Short Papers}, pages 57--60, New York City,
  USA. Association for Computational Linguistics.

\bibitem[{Joshi et~al.(2020)Joshi, Chen, Liu, Weld, Zettlemoyer, and
  Levy}]{joshi-etal-2020-spanbert}
Mandar Joshi, Danqi Chen, Yinhan Liu, Daniel~S. Weld, Luke Zettlemoyer, and
  Omer Levy. 2020.
\newblock \href {https://doi.org/10.1162/tacl_a_00300} {{S}pan{BERT}: Improving
  pre-training by representing and predicting spans}.
\newblock \emph{Transactions of the Association for Computational Linguistics},
  8:64--77.

\bibitem[{Joshi et~al.(2019)Joshi, Levy, Zettlemoyer, and
  Weld}]{joshi-etal-2019-bert}
Mandar Joshi, Omer Levy, Luke Zettlemoyer, and Daniel Weld. 2019.
\newblock \href {https://doi.org/10.18653/v1/D19-1588} {{BERT} for coreference
  resolution: Baselines and analysis}.
\newblock In \emph{Proceedings of the 2019 Conference on Empirical Methods in
  Natural Language Processing and the 9th International Joint Conference on
  Natural Language Processing (EMNLP-IJCNLP)}, pages 5803--5808, Hong Kong,
  China. Association for Computational Linguistics.

\bibitem[{Kantor and Globerson(2019)}]{kantor-globerson-2019-coreference}
Ben Kantor and Amir Globerson. 2019.
\newblock \href {https://doi.org/10.18653/v1/P19-1066} {Coreference resolution
  with entity equalization}.
\newblock In \emph{Proceedings of the 57th Annual Meeting of the Association
  for Computational Linguistics}, pages 673--677, Florence, Italy. Association
  for Computational Linguistics.

\bibitem[{Khosla et~al.(2021)Khosla, Yu, Manuvinakurike, Ng, Poesio, Strube,
  and Ros{\'e}}]{khosla-etal-2021-codi}
Sopan Khosla, Juntao Yu, Ramesh Manuvinakurike, Vincent Ng, Massimo Poesio,
  Michael Strube, and Carolyn Ros{\'e}. 2021.
\newblock \href {https://doi.org/10.18653/v1/2021.codi-sharedtask.1} {The
  {CODI}-{CRAC} 2021 shared task on anaphora, bridging, and discourse deixis in
  dialogue}.
\newblock In \emph{Proceedings of the CODI-CRAC 2021 Shared Task on Anaphora,
  Bridging, and Discourse Deixis in Dialogue}, pages 1--15, Punta Cana,
  Dominican Republic. Association for Computational Linguistics.

\bibitem[{Kobayashi et~al.(2021)Kobayashi, Li, and
  Ng}]{kobayashi-etal-2021-neural}
Hideo Kobayashi, Shengjie Li, and Vincent Ng. 2021.
\newblock \href {https://doi.org/10.18653/v1/2021.codi-sharedtask.2} {Neural
  anaphora resolution in dialogue}.
\newblock In \emph{Proceedings of the CODI-CRAC 2021 Shared Task on Anaphora,
  Bridging, and Discourse Deixis in Dialogue}, pages 16--31, Punta Cana,
  Dominican Republic. Association for Computational Linguistics.

\bibitem[{Lee et~al.(2018)Lee, He, and Zettlemoyer}]{lee-etal-2018-higher}
Kenton Lee, Luheng He, and Luke Zettlemoyer. 2018.
\newblock \href {https://doi.org/10.18653/v1/N18-2108} {Higher-order
  coreference resolution with coarse-to-fine inference}.
\newblock In \emph{Proceedings of the 2018 Conference of the North {A}merican
  Chapter of the Association for Computational Linguistics: Human Language
  Technologies, Volume 2 (Short Papers)}, pages 687--692, New Orleans,
  Louisiana. Association for Computational Linguistics.

\bibitem[{Li et~al.(2021)Li, Kobayashi, and Ng}]{li-etal-2021-codi}
Shengjie Li, Hideo Kobayashi, and Vincent Ng. 2021.
\newblock \href {https://doi.org/10.18653/v1/2021.codi-sharedtask.8} {The
  {CODI}-{CRAC} 2021 shared task on anaphora, bridging, and discourse deixis
  resolution in dialogue: A cross-team analysis}.
\newblock In \emph{Proceedings of the CODI-CRAC 2021 Shared Task on Anaphora,
  Bridging, and Discourse Deixis in Dialogue}, pages 71--95, Punta Cana,
  Dominican Republic. Association for Computational Linguistics.

\bibitem[{Li et~al.(2022)Li, Kobayashi, and Ng}]{li-etal-2022-neural-anaphora}
Shengjie Li, Hideo Kobayashi, and Vincent Ng. 2022.
\newblock \href {https://aclanthology.org/2022.codi-crac.4} {Neural anaphora
  resolution in dialogue revisited}.
\newblock In \emph{Proceedings of the CODI-CRAC 2022 Shared Task on Anaphora,
  Bridging, and Discourse Deixis in Dialogue}, pages 32--47, Gyeongju, Republic
  of Korea. Association for Computational Linguistics.

\bibitem[{Luo(2005)}]{luo-2005-coreference}
Xiaoqiang Luo. 2005.
\newblock \href {https://aclanthology.org/H05-1004} {On coreference resolution
  performance metrics}.
\newblock In \emph{Proceedings of Human Language Technology Conference and
  Conference on Empirical Methods in Natural Language Processing}, pages
  25--32, Vancouver, British Columbia, Canada. Association for Computational
  Linguistics.

\bibitem[{Marasovi{\'c} et~al.(2017)Marasovi{\'c}, Born, Opitz, and
  Frank}]{marasovic-etal-2017-mention}
Ana Marasovi{\'c}, Leo Born, Juri Opitz, and Anette Frank. 2017.
\newblock \href {https://doi.org/10.18653/v1/D17-1021} {A mention-ranking model
  for abstract anaphora resolution}.
\newblock In \emph{Proceedings of the 2017 Conference on Empirical Methods in
  Natural Language Processing}, pages 221--232, Copenhagen, Denmark.
  Association for Computational Linguistics.

\bibitem[{McCowan et~al.(2005)McCowan, Carletta, Kraaij, Ashby, Bourban, Flynn,
  Guillemot, Hain, Kadlec, Karaiskos, Kronenthal, Lathoud, Lincoln, Lisowska,
  Post, Reidsma, and Wellner}]{McCowan2005TheAM}
Iain McCowan, Jean Carletta, Wessel Kraaij, Simone Ashby, Sebastien Bourban,
  Mike Flynn, Ma{\"e}l Guillemot, Thomas Hain, Jaroslav Kadlec, Vasilis
  Karaiskos, Melissa Kronenthal, Guillaume Lathoud, Mike Lincoln, Agnes
  Lisowska, Wilfried Post, Dennis Reidsma, and Pierre~D. Wellner. 2005.
\newblock The {AMI} meeting corpus.
\newblock The AMI Project Consortium, \url{www.amiproject.org}.

\bibitem[{M{\"u}ller(2008)}]{Mller2008FullyAR}
Christoph M{\"u}ller. 2008.
\newblock \emph{Fully Automatic Resolution of ``it'', ``this'', and ``that'' in
  Unrestricted Multi-Party Dialog}.
\newblock Ph.D. thesis, Universit{\"a}t T{\"u}bingen, T{\"u}bingen, Germany.

\bibitem[{Navarretta(2000)}]{navarretta-2000-abstract}
Costanza Navarretta. 2000.
\newblock \href {https://doi.org/10.3115/1117736.1117743} {Abstract anaphora
  resolution in {D}anish}.
\newblock In \emph{1st {SIG}dial Workshop on Discourse and Dialogue}, pages
  56--65, Hong Kong, China. Association for Computational Linguistics.

\bibitem[{Noreen(1989)}]{noreen:book}
Eric~W. Noreen. 1989.
\newblock \emph{Computer-Intensive Methods for Testing Hypotheses: An
  Introduction}.
\newblock John Wiley \& Sons Inc.

\bibitem[{Poesio and Artstein(2008)}]{poesio-artstein-2008-anaphoric}
Massimo Poesio and Ron Artstein. 2008.
\newblock \href
  {http://www.lrec-conf.org/proceedings/lrec2008/pdf/297_paper.pdf} {Anaphoric
  annotation in the {ARRAU} corpus}.
\newblock In \emph{Proceedings of the Sixth International Conference on
  Language Resources and Evaluation ({LREC}'08)}, Marrakech, Morocco. European
  Language Resources Association (ELRA).

\bibitem[{Strube and M{\"u}ller(2003)}]{strube-muller-2003-machine}
Michael Strube and Christoph M{\"u}ller. 2003.
\newblock \href {https://doi.org/10.3115/1075096.1075118} {A machine learning
  approach to pronoun resolution in spoken dialogue}.
\newblock In \emph{Proceedings of the 41st Annual Meeting of the Association
  for Computational Linguistics}, pages 168--175, Sapporo, Japan. Association
  for Computational Linguistics.

\bibitem[{Urbanek et~al.(2019)Urbanek, Fan, Karamcheti, Jain, Humeau, Dinan,
  Rockt{\"a}schel, Kiela, Szlam, and Weston}]{urbanek-etal-2019-learning}
Jack Urbanek, Angela Fan, Siddharth Karamcheti, Saachi Jain, Samuel Humeau,
  Emily Dinan, Tim Rockt{\"a}schel, Douwe Kiela, Arthur Szlam, and Jason
  Weston. 2019.
\newblock \href {https://doi.org/10.18653/v1/D19-1062} {Learning to speak and
  act in a fantasy text adventure game}.
\newblock In \emph{Proceedings of the 2019 Conference on Empirical Methods in
  Natural Language Processing and the 9th International Joint Conference on
  Natural Language Processing (EMNLP-IJCNLP)}, pages 673--683, Hong Kong,
  China. Association for Computational Linguistics.

\bibitem[{Vilain et~al.(1995)Vilain, Burger, Aberdeen, Connolly, and
  Hirschman}]{vilain-etal-1995-model}
Marc Vilain, John Burger, John Aberdeen, Dennis Connolly, and Lynette
  Hirschman. 1995.
\newblock \href {https://aclanthology.org/M95-1005} {A model-theoretic
  coreference scoring scheme}.
\newblock In \emph{Sixth Message Understanding Conference ({MUC}-6):
  Proceedings of a Conference Held in {C}olumbia, {M}aryland, November 6-8,
  1995}.

\bibitem[{Wang et~al.(2019)Wang, Shi, Kim, Oh, Yang, Zhang, and
  Yu}]{wang-etal-2019-persuasion}
Xuewei Wang, Weiyan Shi, Richard Kim, Yoojung Oh, Sijia Yang, Jingwen Zhang,
  and Zhou Yu. 2019.
\newblock \href {https://doi.org/10.18653/v1/P19-1566} {Persuasion for good:
  Towards a personalized persuasive dialogue system for social good}.
\newblock In \emph{Proceedings of the 57th Annual Meeting of the Association
  for Computational Linguistics}, pages 5635--5649, Florence, Italy.
  Association for Computational Linguistics.

\bibitem[{Webber(1991)}]{webber:1991}
Bonnie~Lynn Webber. 1991.
\newblock Structure and ostension in the interpretation of discourse deixis.
\newblock \emph{Language and Cognitive Processes}, 6(2):107--–135.

\bibitem[{Xu and Choi(2020)}]{xu-choi-2020-revealing}
Liyan Xu and Jinho~D. Choi. 2020.
\newblock \href {https://doi.org/10.18653/v1/2020.emnlp-main.686} {Revealing
  the myth of higher-order inference in coreference resolution}.
\newblock In \emph{Proceedings of the 2020 Conference on Empirical Methods in
  Natural Language Processing (EMNLP)}, pages 8527--8533, Online. Association
  for Computational Linguistics.

\bibitem[{Yu et~al.(2022{\natexlab{a}})Yu, Khosla, Manuvinakurike, Levin, Ng,
  Poesio, Strube, and Ros{\'e}}]{yu-etal-2022-codi}
Juntao Yu, Sopan Khosla, Ramesh Manuvinakurike, Lori Levin, Vincent Ng, Massimo
  Poesio, Michael Strube, and Carolyn Ros{\'e}. 2022{\natexlab{a}}.
\newblock \href {https://aclanthology.org/2022.codi-crac.1} {The {CODI}-{CRAC}
  2022 shared task on anaphora, bridging, and discourse deixis in dialogue}.
\newblock In \emph{Proceedings of the CODI-CRAC 2022 Shared Task on Anaphora,
  Bridging, and Discourse Deixis in Dialogue}, pages 1--14, Gyeongju, Republic
  of Korea. Association for Computational Linguistics.

\bibitem[{Yu et~al.(2022{\natexlab{b}})Yu, Khosla, Moosavi, Paun, Pradhan, and
  Poesio}]{yu-etal-2022-universal}
Juntao Yu, Sopan Khosla, Nafise~Sadat Moosavi, Silviu Paun, Sameer Pradhan, and
  Massimo Poesio. 2022{\natexlab{b}}.
\newblock \href {https://aclanthology.org/2022.lrec-1.521} {The universal
  anaphora scorer}.
\newblock In \emph{Proceedings of the Thirteenth Language Resources and
  Evaluation Conference}, pages 4873--4883, Marseille, France. European
  Language Resources Association.

\end{thebibliography}
\bibliographystyle{acl_natbib}

\appendix

\begin{table*}[h!]
\centering
\small
\begin{tabular}{@{}lcccccccccc@{}}
\toprule
\multicolumn{1}{c}{} & \multicolumn{3}{c}{MUC} & \multicolumn{3}{c}{B$^3$} & \multicolumn{3}{c}{CEAF$_e$} &  \\
\cmidrule(lr){2-4} \cmidrule(lr){5-7} \cmidrule(lr){8-10}  \multicolumn{1}{c}{} & P & R & F & P & R & F & P & R & F & CoNLL \\ \midrule
\multicolumn{11}{c}{LIGHT} \\
UTD\_NLP & 44.6 & 31.3 & 36.8 & 56.2 & 37.0 & 44.6 & 55.3 & 40.5 & 46.7 & 42.7 \\
coref-hoi & 37.2 & 36.3 & 36.7 & 48.9 & 42.0 & 45.2 & 58.2 & 38.5 & 46.3 & 42.7 \\
coref-hoi-utt & 36.5 & 38.8 & 37.6 & 46.7 & 42.3 & 44.4 & 55.3 & 38.0 & 45.0 & 42.3 \\
dd-utt & 52.4 & 41.3 & 46.2 & 62.0 & 41.6 & 49.8 & 69.0 & 37.6 & 48.7 & 48.2 \\ \midrule
\multicolumn{11}{c}{AMI} \\
UTD\_NLP & 45.5 & 21.2 & 28.9 & 52.4 & 29.5 & 37.8 & 44.9 & 35.1 & 39.4 & 35.4 \\
coref-hoi & 21.7 & 30.5 & 25.4 & 28.7 & 36.3 & 32.1 & 39.0 & 31.0 & 34.6 & 30.7 \\
coref-hoi-utt & 25.5 & 33.1 & 28.8 & 34.6 & 39.0 & 36.7 & 43.4 & 36.1 & 39.4 & 35.0 \\
dd-utt & 41.2 & 39.8 & 40.5 & 48.9 & 42.8 & 45.6 & 54.4 & 37.5 & 44.4 & 43.5 \\ \midrule
\multicolumn{11}{c}{Persuasion} \\
UTD\_NLP & 45.5 & 20.3 & 28.1 & 65.0 & 30.2 & 41.2 & 61.0 & 41.8 & 49.6 & 39.6 \\
coref-hoi & 48.6 & 42.3 & 45.2 & 57.5 & 45.9 & 51.1 & 66.2 & 44.0 & 52.9 & 49.7 \\
coref-hoi-utt & 50.0 & 49.6 & 49.8 & 56.8 & 51.7 & 54.1 & 64.4 & 49.4 & 55.9 & 53.3 \\
dd-utt & 56.7 & 48.0 & 52.0 & 63.8 & 49.9 & 56.0 & 72.1 & 46.9 & 56.8 & 54.9 \\ \midrule
\multicolumn{11}{c}{Switchboard} \\
UTD\_NLP & 35.2 & 21.3 & 26.5 & 52.3 & 30.4 & 38.5 & 50.5 & 34.9 & 41.3 & 35.4 \\
coref-hoi & 31.5 & 30.4 & 31.0 & 40.9 & 34.0 & 37.1 & 51.4 & 30.2 & 38.0 & 35.4 \\
coref-hoi-utt & 30.6 & 29.3 & 29.9 & 39.5 & 32.7 & 35.8 & 49.5 & 29.2 & 36.7 & 34.1 \\
dd-utt & 46.3 & 43.4 & 44.8 & 54.9 & 44.5 & 49.2 & 63.4 & 38.3 & 47.7 & 47.2 \\ \bottomrule
\end{tabular}
\caption{Resolution results on the test sets.}
\label{tab:DetailedResults}
\end{table*}

\begin{table*}[h!]
\centering
\begin{small}
\begin{tabular}{@{}llcccccccccccc@{}}
\toprule
 &  & \multicolumn{3}{c}{LIGHT} & \multicolumn{3}{c}{AMI} & \multicolumn{3}{c}{Persuasion} & \multicolumn{3}{c}{Switchboard} \\
\cmidrule(lr){3-5} \cmidrule(lr){6-8} \cmidrule(lr){9-11} \cmidrule(l){12-14} &  & P & R & F & P & R & F & P & R & F & P & R & F \\ \midrule
\multirow{4}{*}{Overall} & UTD\_NLP & 65.2 & 46.9 & 54.6 & 60.2 & 39.1 & 47.4 & 72.3 & 41.6 & 52.8 & 64.4 & 42.2 & 51.0 \\
 & coref-hoi & 62.9 & 49.5 & 55.4 & 40.5 & 42.7 & 41.5 & 68.6 & 52.0 & 59.2 & 55.3 & 41.2 & 47.2 \\
 & coref-hoi-utt & 59.3 & 50.0 & 54.2 & 43.9 & 45.2 & 44.5 & 66.2 & 57.6 & 61.6 & 53.3 & 39.6 & 45.5 \\
 & dd-utt & 72.6 & 46.9 & 57.0 & 57.8 & 46.6 & 51.6 & 73.9 & 54.7 & 62.8 & 66.9 & 49.6 & 57.0 \\ \midrule
\multirow{4}{*}{Anaphor} & UTD\_NLP & 71.4 & 68.8 & 70.1 & 58.0 & 64.4 & 61.0 & 76.7 & 64.2 & 69.9 & 65.7 & 70.7 & 68.1 \\
 & coref-hoi & 71.8 & 70.0 & 70.9 & 42.2 & 59.3 & 49.3 & 72.9 & 63.4 & 67.8 & 63.0 & 60.8 & 61.9 \\
 & coref-hoi-utt & 68.2 & 72.5 & 70.3 & 46.4 & 60.2 & 52.4 & 71.3 & 70.7 & 71.0 & 61.9 & 59.3 & 60.6 \\
 & dd-utt & 81.0 & 63.8 & 71.3 & 57.9 & 55.9 & 56.9 & 77.9 & 65.9 & 71.4 & 67.5 & 63.1 & 65.2 \\ \midrule
\multirow{4}{*}{Antecedent} & UTD\_NLP & 50.8 & 27.7 & 35.8 & 66.0 & 20.5 & 31.3 & 59.6 & 21.2 & 31.3 & 60.8 & 21.5 & 31.7 \\
 & coref-hoi & 52.7 & 34.8 & 41.9 & 38.3 & 30.4 & 33.9 & 63.9 & 42.5 & 51.0 & 46.3 & 27.2 & 34.3 \\
 & coref-hoi-utt & 49.4 & 33.9 & 40.2 & 41.0 & 34.2 & 37.3 & 60.7 & 46.6 & 52.7 & 43.3 & 25.5 & 32.1 \\
 & dd-utt & 63.9 & 34.8 & 45.1 & 57.7 & 39.8 & 47.1 & 69.5 & 45.2 & 54.8 & 66.2 & 40.0 & 49.8 \\ \bottomrule
\end{tabular}
\end{small}
\caption{Mention extraction results on the test sets.}
\label{tab:MenExtract}
\end{table*}

\section{Detailed Experimental Results}
\label{app:DetailedScores}

We report the resolution results of the four resolvers ({\tt UTD\_NLP}, {\tt coref-hoi}, {\tt coref-hoi-utt}, and {\tt dd-utt}) on the CODI-CRAC 2021 shared task test sets in terms of
MUC, B$^3$, and CEAF$_e$ scores in Table \ref{tab:DetailedResults} and their mention extraction results in terms of recall (R), precision (P), and F-score (F) in Table \ref{tab:MenExtract}.

Consider first the resolution results in Table \ref{tab:DetailedResults}. As can be seen, not only does {\tt dd-utt} achieve the best CoNLL scores on all four datasets, but it does so via achieving the best MUC, B$^3$, and CEAF$_e$ F-scores. In terms of MUC F-score, the performance difference between {\tt dd-utt} and the second best resolver on each dataset is substantial (2.2\%--14.9\% points). These results suggest that better link identification, which is what the MUC F-score reveals, is the primary reason for the superior performance of {\tt dd-utt}. Moreover, Persuasion appears to be the easiest of the four datasets, as this is the dataset on which three of the four resolvers achieved the highest CoNLL scores. Note that Persuasion is also the dataset on which
the differences in CoNLL score between {\tt dd-utt} and the other resolvers are the smallest. These results seem to suggest that the performance gap between {\tt dd-utt} and the other resolvers tends to widen as the difficulty of a dataset increases.

Next, consider the {\em anaphor} extraction results in Table \ref{tab:MenExtract}. In terms of F-score, {\tt dd-utt} lags behind {\tt UTD\_NLP} on two datasets, AMI and Switchboard. Nevertheless, the anaphor extraction {\em precision} achieved by {\tt dd-utt} is often one of the highest in each dataset.

\end{document}